\documentclass[journal]{IEEEtran}
\usepackage{times}  
\usepackage{helvet}  
\usepackage{courier}  
\usepackage{url}  
\usepackage{graphicx}  
\usepackage{epstopdf}
\usepackage{amsmath,amssymb} 
\usepackage{color}
\usepackage{cite}
\usepackage{booktabs}
\usepackage{threeparttable}
\usepackage{times}
\usepackage{xcolor}
\usepackage{subfigure}
\usepackage{slashbox}
\usepackage{float}
\usepackage{multirow}
\usepackage{multicol}
\usepackage{amsmath}
\usepackage{color}
\usepackage{soul}
\usepackage[utf8]{inputenc}
\usepackage[small]{caption}
\usepackage{float}
\usepackage[colorlinks,
            linkcolor=red,
            anchorcolor=blue,
            citecolor=green,
            backref=page
            ]{hyperref}
\usepackage{cleveref}

\def\x{{\mathbf x}}

\begin{document}

\title{PSGCNet: A Pyramidal Scale and Global Context Guided Network for Dense Object Counting in Remote Sensing Images}

\author{Guangshuai Gao,
        Qingjie Liu,~\IEEEmembership{Member,~IEEE},
        Zhenghui Hu,
        Lu Li,
        Qi Wen,
        Yunhong Wang,~\IEEEmembership{Fellow,~IEEE}}

\maketitle

\begin{abstract}
Object counting, which aims to count the accurate number of object instances in images, has been attracting more and more attention. However, challenges such as large scale variation, complex background interference, and non-uniform density distribution greatly limit the counting accuracy, particularly striking in remote sensing imagery. To mitigate the above issues, this paper proposes a novel framework for dense object counting in remote sensing images, which incorporates a pyramidal scale module (PSM) and a global context module (GCM), dubbed PSGCNet, where PSM is used to adaptively capture multi-scale information and GCM is to guide the model to select suitable scales generated from PSM. Moreover, a reliable supervision manner improved from Bayesian and Counting loss (BCL) is utilized to learn the density probability and then compute the count expectation at each annotation. It can relieve non-uniform density distribution to a certain extent. Extensive experiments on four remote sensing counting datasets demonstrate the effectiveness of the proposed method and the superiority of it compared with state-of-the-arts. Additionally, experiments extended on four commonly used crowd counting datasets further validate the generalization ability of the model. Code is available at \url{https://github.com/gaoguangshuai/PSGCNet}.
\end{abstract}


\IEEEpeerreviewmaketitle

\section{Introduction}
\IEEEPARstart{O}{bject} counting, which is to estimate the accurate number of object instances in images or videos, has been attracting remarkable interest in recent years owing to its potential value in traffic monitor~\cite{kang2018beyond}, urban planning~\cite{li2014crowded}, public safety~\cite{zhang2017understanding}, and crowd behavior understanding~\cite{zhang2016data}, etc. Additionally, object counting has been applied in many practical applications, such as cell microscopy~\cite{lempitsky2010learning}, animals~\cite{arteta2016counting}, and remote sensing applications~\cite{gao2020counting,gao2020counting_2,hsieh2017drone,du2018unmanned}.

\begin{figure}[t]
	\centering
	\includegraphics[width=1.0 \linewidth]{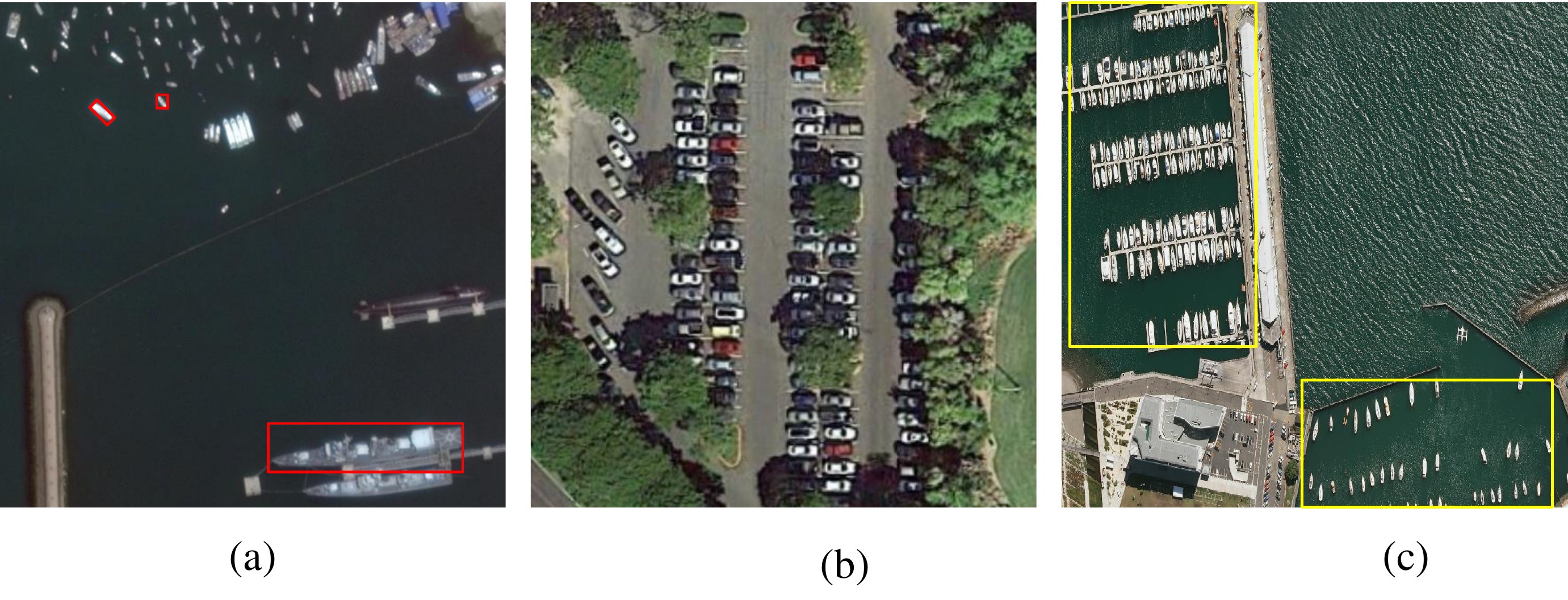}
	\caption{{Illustrations of large scale variation, complex background interference and non-uniform density distribution. In (a), scales of ships enclosed by the red bounding boxes vary largely. In (b), the objects (i.e., small vehicles) are shaded by the plants. In (c), ships at harbor are unevenly distributed.}}
	\label{fig:sample}
\end{figure}

Recent prevalent object counting methods have been following the pioneering work~\cite{lempitsky2010learning}, which estimates the count number over a density map. Lately, driven by the powerful feature representation ability of Convolutional Neural Networks (CNNs), a lot of CNN-based density estimation algorithms have been presented. Although remarkable progress has been achieved, there still exist challenges limiting the counting performance, such as large scale variation, complex background interference, non-uniform density distribution, which are much tougher in remote sensing images. Taking Fig.\ref{fig:sample} (a) as an example, the scale variation of different ship instances is large because of different types of ships. In Fig.\ref{fig:sample} (b), complex background interferences (such as the green plants) are easy to fool models to make wrong predictions. Furthermore, in Fig.\ref{fig:sample} (c), the spatial distribution is non-uniform varying from sparse to congested even in the same scene.

Many efforts tackle the scale variation problem by designing multi-column architectures~\cite{zhang2016single,sindagi2017generating,cheng2019improving} or employing techniques, such as dilated convolution~\cite{yu2016multi,lan2020global}, Spatial Pyramid Pooling (SPP)~\cite{he2015spatial,lan2020global}, Atrous Spatial Pyramid Pooling (ASPP)~\cite{chen2017deeplab,zhang2021stagewise}, and Inception blocks~\cite{szegedy2015going} to capture multi-scale information~\cite{li2018csrnet,chen2019scale,liu2019context-aware,cao2018scale}. These models relieve scale variation problem yet still have some limitations. 1) Multi-column architectures or Inception blocks have multiple branches built with different kernel sizes, which introduce a large number of parameters and huge computation burdens~\cite{he2019dynamic}. 2) The pooling operation in these models (e.g., SPP) may lead to fine-detail information loss, thus degenerating the performance. 3) Hand-crafted dilation rates are hard to match the range of scale variations.
To alleviate these issues, motivated by \cite{duta2020pyramidal}, we embed a pyramidal scale module (PSM) into our framework to effectively capture multi-scale information.

To suppress background distractors, visual attention has been successfully applied to the counting task~\cite{liu2019adcrowdnet,gao2020counting,gao2020counting_2} and achieves good performance. However, these attention modules suffer from heavy computation cost and high complexity, for instance, the fashionable Squeeze-and-Excitation network (SENet)~\cite{hu2018squeeze} and its followers~\cite{woo2018cbam,li2019selective} employ multiple fully connected (FC) layers to compute attention weights. Such designs are inefficient and not helpful for capturing the interactive information across channels. Inspired by \cite{wang2020eca,yang2020gated}, we introduce an effective and efficient global context module (GCM) to select more suitable scales generated from PSM.

Most counting methods convert annotated points into density maps using Gaussian filters and then train CNN models using $L_2$ loss. Consequently, the counting performance highly depends on the quality of the ``ground truth" density map. However, such a pixel-independent based ``ground-truth" density map generation manner may be suboptimal, especially in non-uniform distributed regions. As an alternation, Ma et al.~\cite{ma2019bayesian} propose a reliable supervision manner through learning the count expectation from the point annotations, named Bayesian Loss (BL). This effective supervision manner could alleviate the problem of non-uniform density distribution. However, there may exist an inconsistency between the training phase (point-to-point loss) and the testing stage (the difference between the overall summation of estimated density maps and ground truth counts). Therefore, apart from the Bayesian loss, we add a counting loss to mitigate this issue.

In summary, the contributions of this work are three-fold:

\begin{itemize}
  \item A novel \textbf{P}yramidal \textbf{S}cale and \textbf{G}lobal \textbf{C}ontext-based framework for dense object counting in remote sensing images, termed PSGCNet, is presented. 
  \item A flexible pyramid scale module is designed to effectively extract multi-scale features of dense scenes. And a lightweight global context module is embedded to make use of the rich interaction information across channels of feature map to guide the model to select more suitable scales.
  \item Extensive experiments conducted on four remote sensing object counting datasets demonstrate the effectiveness and superiority of the proposed approach, and the extension to four commonly used crowd counting datasets further validate the generalization ability and robustness of our proposed method.
\end{itemize}

\begin{figure*}[t]
	\centering
	\includegraphics[width=1.0 \linewidth]{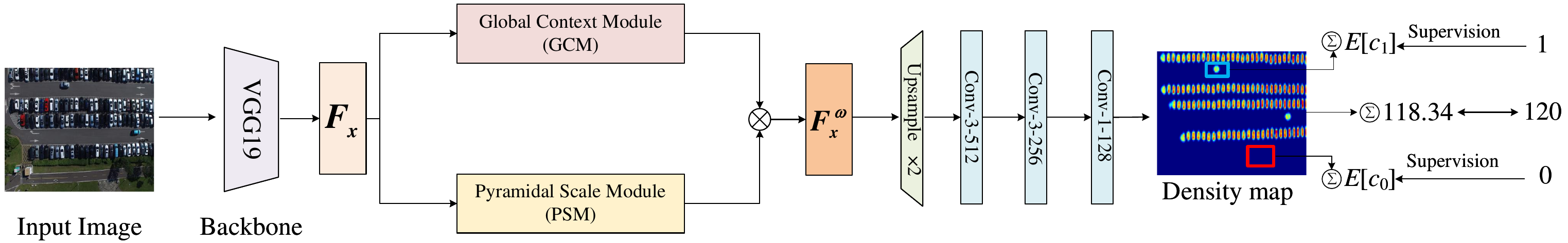}
	\caption{The architecture of PSGCNet for object counting in remote sensing images. The parameters of the convolution layers are denoted as ``Conv-(kernel\_size)-(number of filters)". ``$\otimes$" indicates the element multiplication operation. We take VGG19 as the backbone to extract features $F_x$ from an input image, which further pass through the proposed GCM and PSM modules to generate enhanced features $F_x^{\omega}$. The predicted density map is produced after one upsampling and three convolution layers. We train the whole pipeline with a hybrid loss by combining Bayesian and counting loss. The red and blue rectangles are the background pixels and activations of an object.}
	\label{fig:architecture}
\end{figure*}

The remainder of this paper is organized as follows. The related work of object counting algorithms is briefly surveyed in Section \ref{sec:related}. The details of our proposed method are introduced in Section \ref{sec:method}, following which experimental results and analysis are presented in Section \ref{sec:experimets}. Finally, the conclusion is concluded in Section \ref{sec:conclusion}.
\section{Related Work}
\label{sec:related}

\subsection{Object counting in congested scenes}
Early object counting methods are mainly detection-based~\cite{kamenetsky2015aerial,moranduzzo2013automatic}, they first detect the interested object instances and then count the number of the bounding boxes. These methods obtain satisfactory performance in sparse scenarios thanks to the powerful detectors. However, they may fail in highly congested scenes, since the object instances are usually with small sizes and easily confused with background distractors. Another mainline is regression-based methods, which map the high dimension image space to natural numbers~\cite{an2007face,chan2008privacy}. As a highly non-linear regression task, it is very hard to optimize models and the performance is far from satisfactory. \cite{lempitsky2010learning} rekindles the counting task as a density map generation problem, which estimates the counting number of object instances by integrating all the pixels of the density map. Entering the deep learning era, the performance of object counting has been significantly improved. Many deep neural networks have been designed for tackling the counting task. The performances on several representative benchmark datasets such as ShanghaiTech~\cite{zhang2016single}, UCF-QNRF~\cite{idrees2018composition}, and UCF\_CC\_50~\cite{idrees2013multi} have reached promising results. For a comprehensive review of the counting task, please refer to~\cite{gao2020cnn,sindagi2018survey}.
\subsection{Object counting from the remote sensing view}
Capturing from a remote distance, aerial images or videos provide a wider field of view and thus with much more complex scene contents, which brings great challenges to existing counting models. To facilitate research in this field, \cite{bahmanyar2019mrcnet} introduces a drone-based crowd dataset and develops a multi-resolution network for estimating the number of pedestrians in aerial images. LPN~\cite{hsieh2017drone} takes advantage of the regular spatial layout of cars and proposes a spatial layout proposal network for car counting and localization, simultaneously. LEP~\cite{stahl2018divide} proposes to predict the image-level count by dividing the image into a set of divisions. It achieves good performance on several drone-based counting datasets. Li et al.~\cite{li2019simultaneously} draws inspiration from object detectors and proposes to detect and count cars simultaneously using a unified framework. STNNet~\cite{wen2021detection} takes a step further and performs the density map estimation, localization, and tracking tasks in one network. ASPDNet~\cite{gao2020counting_2} builds a new benchmark for aerial image counting. It employs recently developed techniques such as dilated convolution, attention, and deformable convolution to achieve a better performance.
\subsection{Alleviating large scale variation}
Scale variation is a great challenge for object counting. Four strategies are widely studied to address this problem: multi-column network architectures, dilated convolution, Spatial Pyramid Pooling (SPP), and Inception module. For example, MCNN~\cite{zhang2016single} is a simple multi-column network, in which, each column is built with different filter kernels. Switch-CNN~\cite{sam2017switching} adopts a frame structure similar to MCNN~\cite{zhang2016single}. The difference is that a specialized classifier is applied to select a suitable column network for inputs.
CSRNet~\cite{li2018csrnet} takes advantage of dilated convolution to enlarge the receptive fields without increasing computation cost.
CAN~\cite{liu2019context-aware} combines scale-aware and context-aware feature information to boost the performance. SANet~\cite{cao2018scale} captures multi-scale features built on the shoulder of the Inception module~\cite{szegedy2015going}. DSNet~\cite{dai2021dense} cascades multiple dense dilated convolution blocks and links them with dense residual connections.
ADSCNet~\cite{bai2020adaptive} adopts adaptive dilated convolution to learn dynamic and continuous dilated rates for each pixel location. MRCNet~\cite{bahmanyar2019mrcnet} combines low-level and high-level features with lateral connections to learn contextual and detailed local information in aerial imagery. SACANet~\cite{bai2019crowd} utilizes a pyramid contextual module to extract long-range contextual information and enlarge the receptive fields of the objects in drone scenes. ASPDNet~\cite{gao2020counting,gao2020counting_2} integrates a scale pyramid module to capture multi-scale information for counting in remote sensing images.

\subsection{Mitigating cluttered background interferences}
Attention mechanism has been widely used to suppress cluttered backgrounds and highlight foreground regions. For instance, SAANet~\cite{varior2019scale} develops a soft attention mechanism to learn a set of gating masks to aggregate the multi-scale density maps. ADCrowdNet~\cite{liu2019adcrowdnet} combines visual attention and deformable convolution~\cite{dai2017deformable} into a unified framework. HA-CNN~\cite{sindagi2019ha} designs a hierarchical attention based network to selectively enhance the features at various levels. RANet~\cite{zhang2019relational} and ANF~\cite{zhang2019attentional} incorporate self-attention to capture long-range dependencies of the feature maps. SDANet~\cite{miao2020shallow} builds a dense attention network based on shallow features. ASNet~\cite{jiang2020attention} learns attention scaling factors and automatically adjusts the density regions by multiplying multiple density attention masks on them. SACANet~\cite{bai2019crowd} leverages a scale-adaptive self-attention multi-branch module to address isolated clusters in aerial images. ASPDNet~\cite{gao2020counting,gao2020counting_2} cascades channel attention and spatial attention to relieve the impact of complex cluttered backgrounds in diverse remote sensing scenarios. These methods have gained significant performance, nevertheless, the sophisticated structures of the attention modules incorporated in them introduce a large number of parameters, thus making them suffer from huge computation burdens. Although some lightweight attention modules such as Squeeze-and-Excitation networks (SENet)~\cite{hu2018squeeze} and convolution block attention module (CBAM)~\cite{woo2018cbam} are developed to alleviate this problem, the fully connected (FC) layers still have many parameters. What's more, the channel dimensionality reduction in these models also limits the upper bound of the performance.

Different from the aforementioned methods, our proposed PSGCNet takes advantage of a pyramidal scale module to capture multi-scale features, which can flexibly cover various scales and enlarge the receptive field without increasing any computation cost. Additionally, we devise an effective global context module, essentially a lightweight channel attention operation. It can not only reduce the computation burden of attention modules, but also make the cross-channel interaction more efficient by avoiding dimensionality reduction. Finally, we train our model with a reliable supervision manner on the count expectation at each annotation point.

\section{Proposed Method}
\label{sec:method}
\subsection{PSGCNet Overview}
The architecture of PSGCNet is illustrated in Fig.~\ref{fig:architecture}. It has four key components, including a backbone network as feature extractor, a pyramidal scale module capturing multi-scale information, a global context module suppressing cluttered backgrounds, and a decoder to estimate the final density map.

Specifically, we adopt a truncated VGG19~\cite{simonyan2015very} same to \cite{ma2019bayesian} as the backbone network, in which the three fully connected layers and one pooling layer are removed. The output feature map's resolution of the backbone is 1/16 of the original input image. Afterwards, a pyramid scale module is built on top of the feature maps to capture multi-scale information. Then, an effective global context module (GCM) followed is leveraged to restrain the complex backgrounds. Then feature maps are upsampled twice with bilinear interpolation operation. Finally, a decoder is equipped to produce the density map, in which three successive convolutional layers are used, including two 3$\times$3 convolution layers with 256 and 128 channels, and one 1$\times$1 convolution. To further improve the performance, we optimize the model using a modified Bayesian Loss.

\subsection{Pyramidal Scale Module (PSM)}
Scale variation is a critical problem in remote sensing image understanding. In this paper, we attack this problem by introducing a pyramidal scale module (PSM). PSM deploys two paralleled network paths: a local PyConv path and a global PyConv path. The two paths have a dual-oriented pyramid architecture, enabling richer multi-scale information capturing.

\begin{figure}[ht!]
	\centering
	\includegraphics[width=1.0 \linewidth]{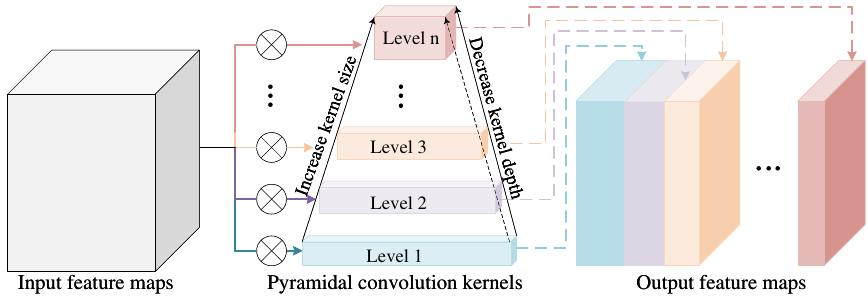}
	\caption{The sketch of pyramidal convolution (PyConv).}
	\label{fig:pykernel}
\end{figure}

PyConv has a pyramid structure, as shown in Fig.~\ref{fig:pykernel}. It contains increasing kernel sizes from bottom to top in a pyramidal manner, and decreasing kernel depths (connectivity) with grouped convolution. The double-oriented pyramid operation allows the model to capture richer multi-scale information, from larger receptive fields of kernels with lower connectivity to smaller receptive fields with higher connectivity. This design is efficient, flexible and economical computational cost.

\begin{figure}[ht!]
	\centering
	\includegraphics[width=0.8 \linewidth]{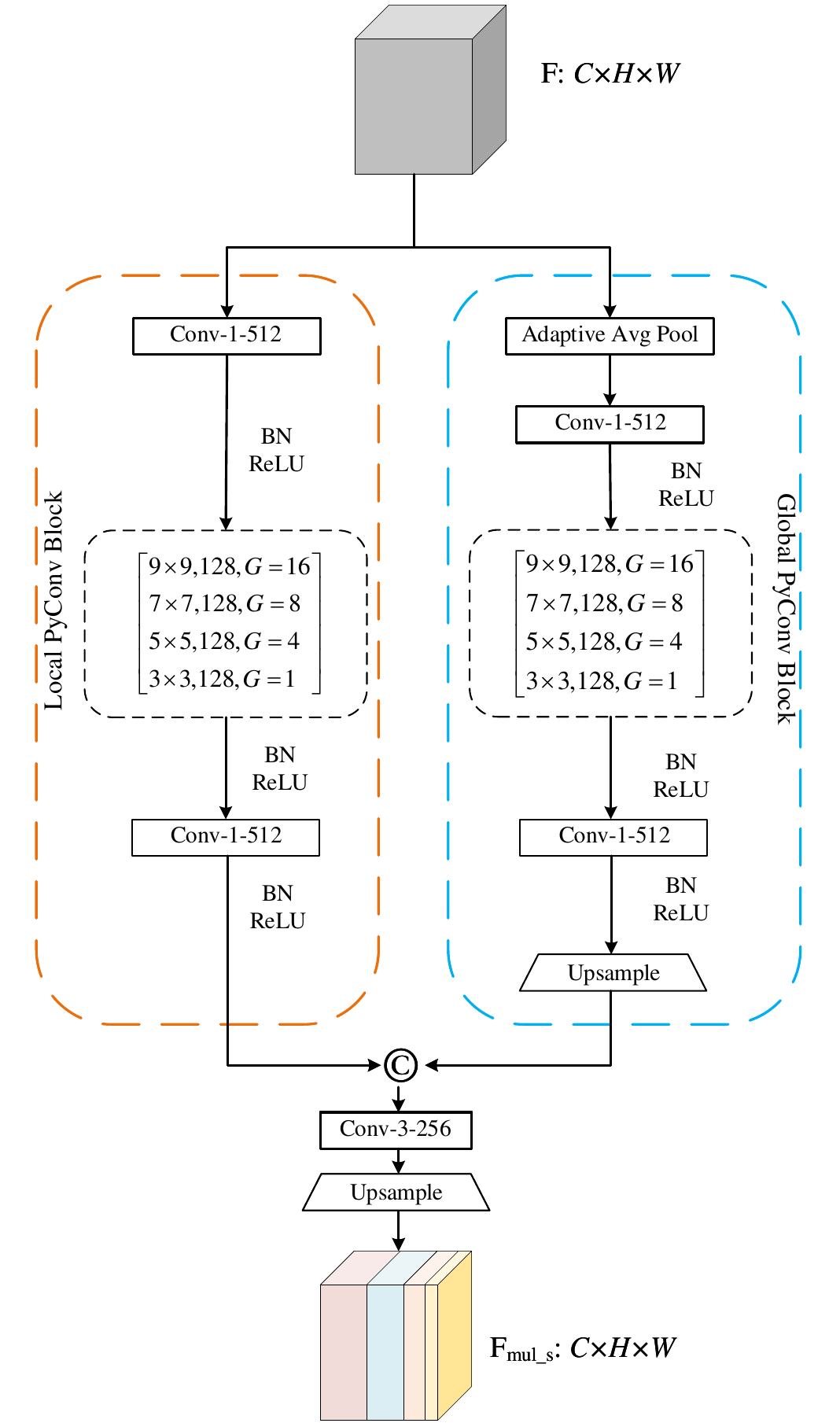}
	\caption{The detail architecture and parameters of PSM. ``$\copyright$'' indicates the concatenation operation.}
	\label{fig:pyramidal}
\end{figure}

The local PyConv path (the left branch in Fig.~\ref{fig:pyramidal}) has smaller receptive fields, which is responsible for tiny objects. It firstly applies 1$\times$1 convolutions to reduce the channels to 512 and then aggregates four layers with different kernel sizes (i.e., 9$\times$9, 7$\times$7, 5$\times$5, and 3$\times$3). Besides, the number of groups (G) enables the kernels to have different connectivity. This is achieved with 1$\times$1 convolutions. Note that each convolution block is followed by a batch normalization layer and a ReLU activation layer.

The global PyConv path (the right branch in Fig.~\ref{fig:pyramidal}) is to capture features of large objects in a global perspective. It has a similar structure to the local PyConv block, however, uses an adaptive average pooling operation on the top to reduce the spatial size of the feature maps to 9$\times$9 and upsample the feature maps to the same resolution to the input through bilinear interpolation at the bottom.

The features from both local and global PyConv blocks are then concatenated and followed by a standard convolution layer with the size of 3$\times$3. Finally, we upsample the feature maps to the original image size. The PSM module is efficient, flexible, and economical in computational cost. It could also boost the robustness of the model to scale variation.
\subsection{Global Context Module (GCM)}
Visual attention has been claimed as a promising solution to overcome the interference of complex backgrounds. These models have achieved improved performance, however, with a cost of higher model complexities and heavier computational burden, since they usually use self-attention~\cite{vaswani2017attention} or non-local modules~\cite{wang2018non}.

\begin{figure}[t]
	\centering
	\includegraphics[width=1.0 \linewidth]{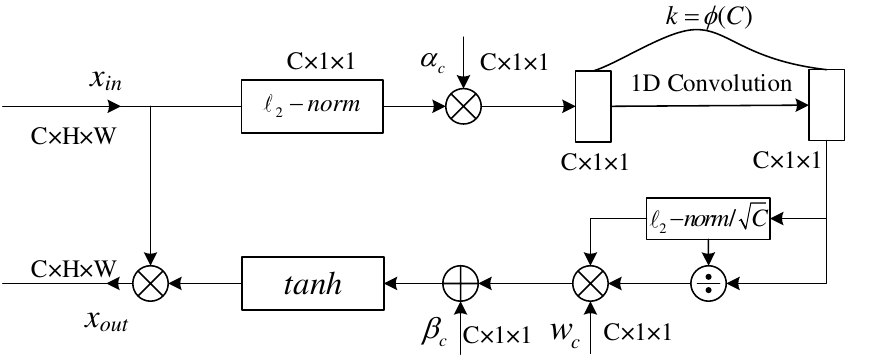}
	\caption{Illustration of the global context module.}
	\label{fig:channel}
\end{figure}

Drawing inspiration from \cite{wang2020eca} and \cite{yang2020gated}, we propose an efficient and lightweight global context module to model the dependencies across the channels. The global context module designed in our work is depicted in Fig.~\ref{fig:channel}.

Concretely, given an intermediate feature map, denoted as $x\in\mathbb{R}^{C\times H \times W}$, where $C$, $H$, and $W$ represent the number of channels, height, and width of the feature map, respectively. Let $x_{c}$ be the feature map corresponding to the \emph{c}-th channel, i.e., $x_{c}=\left[x_{c}^{i, j}\right]_{H \times W} \in \mathbb{R}^{H \times W}, c \in\{1,2, \cdots, C\}$. A global context module is embedded to capture global context information of each channel. The module is formulated as:
\begin{equation}
s_{c}=\alpha_{c}\left\|x_{c}\right\|_{2}=\alpha_{c}\left\{\left[\sum_{i=1}^{H} \sum_{j=1}^{W}\left(x_{c}^{i, j}\right)^{2}\right]+\epsilon\right\}^{\frac{1}{2}}
\end{equation}
where $\alpha_{c}$ denotes the embedding weight, and $\epsilon$ is a small constant to avoid the deviation at zero points. This global context module is somewhat similar to the global average pooling (GAP) but more robust than it~\cite{yang2020gated}.

Generally, to effectively learn cross-channel interactions, typical solutions are SENet~\cite{hu2018squeeze} or CBAM~\cite{woo2018cbam}, however, they destroy the correspondence between channels. Here, we adopt an alternation strategy, which first adaptively determines the kernel sizes $k$ ($k=3$ in this paper) and then performs a 1D convolution operation, i.e.,
\begin{equation}
\hat{s}_{c}=C1D\left(s_{c}\right)
\end{equation}
where $C1D$ means $1D$ convolution.

A subsequent channel normalization is applied, which can be formulated as:
\begin{equation}
\tilde{s}_{c}=\frac{\sqrt{C}\hat{s}_{c}}{\|\mathbf{s}\|_{2}}=\frac{\sqrt{C} \hat{s}_{c}}{\sqrt{\sum_{c=1}^{C} \hat{s}_{c}^{2}+\epsilon}}
\end{equation}

Eventually, the final global context attention map $\tilde{x}_{c}^{a t t} \in \mathbb{R}^{C \times 1 \times 1}$ is obtained after a $tanh$ activation layer:
\begin{equation}
\tilde{x}_{c}^{a t t}=\tanh \left(w_{c} \tilde{s}_{c}+\beta_{c}\right)
\end{equation}
where $w_{c}$ and $\beta_{c}$ represent the trainable weight and bias, which are both initialized to 0 in the training stage.

\subsection{Bayesian and counting loss function (BCL)}
To optimize models, Euclidean distance ($L_2$ loss) between the prediction and the ground truth density maps is widely used. However, the loss is not robust to the occlusion, scale variation, and non-uniform density. Recently, Ma et al.~\cite{ma2019bayesian} propose a novel supervision manner, named Bayesian Loss to relieve this problem. It constructs a density contribution model from point annotations and then defines the loss as the difference between the count expectation and the ground truth number at each annotated point:

\begin{equation}
\mathcal{L}^{\text {Bayesian}}=\sum_{n=1}^{N} \mathcal{F}\left(1-E\left[c_{n}\right]\right)+\mathcal{F}\left(0-E\left[c_{0}\right]\right)
\end{equation}
where $N$ is the total number of labelled objects, $E\left[c_{n}\right]$ and $E\left[c_{0}\right]$ indicate the expected counts for each instance and the entire background, respectively. The first term denotes that impelling the foreground count at each annotation point equals 1, while the second term means enforcing the background count to be zero. $\mathcal{F}(\cdot)$ is a distance function, we adopt $\ell_{1}$ distance metric as suggested in \cite{ma2019bayesian}.

Although reliable and effective, there may exist inconsistency between the training phase and the testing stage. Therefore, apart from Bayesian loss, we add a counting loss to mitigate this issue. The counting loss is defined as:

\begin{equation}
\mathcal{L}^{\text {Count}}=\frac{1}{N} \sum_{i=1}^{N}\left\|F\left(X_{i} ; \Theta\right)-Y_{i}\right\|_{1}
\end{equation}
where $F\left(X_{i} ; \Theta\right)$ and $Y_{i}$ represent the count integrated by the estimated density map and ground truth count of the $i$-th image. $\Theta$ denotes training parameters and $\|\cdot\|_{1}$ means $\ell_{1}$-norm.

Therefore, the overall loss function is the combination of Bayesian loss $\mathcal{L}^{\text {Bayesian}}$ and counting loss $\mathcal{L}^{\text {Count }}$:
\begin{equation}
\mathcal{L}^{\text {Overall}}=\mathcal{L}^{\text {Bayesian}}+\lambda\mathcal{L}^{\text {Count}}
\end{equation}
where $\lambda$ is a tunable positive hyperparameter.
\section{Experimental results}
\label{sec:experimets}
In this section, the datasets, evaluation protocols, and implementation details are first introduced. Then ablation studies and comparisons with state-of-the-art methods are provided to demonstrate the effectiveness and superiority of the proposed approach. Furthermore, some extension experiments to other object counting applications are conducted to validate the generalization ability and robustness of the model.

\begin{table*}[!htb]
	\caption{Statistics of the object counting datasets. Total-, min-, average- and max represent the total number, the minimum, average number and maximum number of instances in the datasets, respectively.}
	\begin{center}
		\resizebox{\textwidth}{!}{
    \begin{tabular}{c|c|c|c|c|c|c|c|c}

    \hline

    \multicolumn{1}{c|}{\multirow {2}{*}{Dataset}} &\multicolumn{1}{c|}{\multirow {2}{*}{Sensor}} &\multicolumn{1}{c|}{\multirow {2}{*}{\#Images}} &\multicolumn{1}{c|}{\multirow {2}{*}{Training/Test}} &\multicolumn{1}{c|}{\multirow{2}{*}{Average Resolution}} &\multicolumn{2}{r}{Count Statistics} \\
    \cline{6-9}
    ~ &~ &~ &~ &~  &Total &Min  &Average &Max \\ \hline
    RSOC\_building~\cite{gao2020counting} &Satellite &2468 &1205/1263 &512$\times$512  &76,215 &15 &30.88 &142 \\
	RSOC\_small-vehicle~\cite{gao2020counting} &Satellite &280 &222/58 &2473$\times$2339  &148,838 &17 &531.56 &8531 \\
	RSOC\_large-vehicle~\cite{gao2020counting} &Satellite &172 &108/64 &1552$\times$1573  &16,594 &12 &96.48 &1336 \\
	RSOC\_ship~\cite{gao2020counting} &Satellite &137 &97/40 &2558$\times$2668  &44,892 &50 &327.68 &1661 \\ \hline
    \hline
    CARPK~\cite{hsieh2017drone} &Drone &1448 &989/459 &720$\times$1280 &89,777 &1 &62 &188 \\ \hline
    PUCPR+~\cite{hsieh2017drone} &Camera &125 &100/25 &720$\times$1280 &16,915 &0 &135 &331 \\ \hline
    DroneCrowd~\cite{wen2021detection} &Drone &33,600 &24,600/9,000 &1920$\times$1080 &4,864,280 &25 &144.8 &455\\ \hline \hline
    SHT\_A~\cite{zhang2016single} &CCTV  & 482 &300/182 & 589 $\times$ 868 & 241,677 & 33 & 501.4 & 3,139 \\
    SHT\_B~\cite{zhang2016single} &CCTV & 716 &400/316 & 768 $\times$ 1024 & 88,488 & 9 & 123.6 & 578 \\
    UCF-QNRF~\cite{idrees2018composition} &CCTV  & 1,535 &1201/334 & 2013 $\times$ 2902 & 1,251,642 & 49 & 815 & 12,865 \\
    UCF\_CC\_50~\cite{idrees2013multi}  &CCTV & 50 &-- & 2101 $\times$ 2888 & 63,974 & 94 & 1,280 & 4,543 \\

    \hline

    \end{tabular}}
    \label{tab:datasets}
    \end{center}
\end{table*}

\subsection{Datasets and evaluation protocols}
\textbf{Datasets:} Extensive experiments are conducted on four remote sensing object counting datasets including RSOC~\cite{gao2020counting,gao2020counting_2}, CARPK~\cite{hsieh2017drone}, PUCPR+~\cite{hsieh2017drone}, and Drone-crowd~\cite{wen2021detection} to evaluate the effectiveness and superiority of the proposed approach. Moreover, to validate the generalization ability and robustness of the model, we also conduct experiments on four widely used crowd counting datasets, i.e., ShanghaiTech Part\_A and Part\_B~\cite{zhang2016single} , UCF-QNRF~\cite{idrees2018composition}, and UCF\_CC\_50~\cite{idrees2013multi}. The statistics of the datasets is presented in Table~\ref{tab:datasets}.

\noindent $\bullet$ \textbf{RSOC}~\cite{gao2020counting,gao2020counting_2}\footnote{https://github.com/gaoguangshuai/Counting-from-Sky-A-Large-scale-Dataset-for-Remote-Sensing-Object-Counting-and-A-Benchmark-Method} is a remote sensing object counting dataset, which is composed of four categories, including buildings, small vehicles, large vehicles, and ships. The dataset consists of 3,057 images with 286,539 instances in total.
In which 2,468 building images, 1,205 and 1,263 are used for training and testing; 280 small vehicle images, 222 images for training and 58 for testing; 172 large vehicle images, 108 for training and 64 for testing; 137 ship images, 97 images for training and 40 images for testing, respectively.

\noindent $\bullet$ \textbf{CARPK}~\cite{hsieh2017drone}\footnote{https://lafi.github.io/LPN/} is a large-scale drone-view car counting dataset, which contains 1,448 images with nearly 90k cars in total, of which 989 images for training and the remaining 459 images for testing.

\noindent $\bullet$ \textbf{PUCPR+}~\cite{hsieh2017drone}\footnote{https://lafi.github.io/LPN/} is also a car counting dataset, all the images are captured from the 10th floor of a building. The dataset contains 125 images with approximately 17k cars, of which 100 images are served as training set, and the rest as testing set.

\noindent $\bullet$ \textbf{Drone-crowd}~\cite{wen2021detection}\footnote{https://github.com/VisDrone/VisDrone-Dataset} is a drone-captured dataset for density map estimation, crowd localization and tracking, which is composed of 112 video clips with 33,600 frames in total. The video clips are annotated with over 4.8 million head annotations and several video-level attributes. All the images are captured by drone-mounted cameras in 70 different scenarios across 4 different cities in China (i.e., Tianjin, Guangzhou, Daqing, and Hong Kong). For the counting task in this paper, we split the dataset into training and test set, of which 24,600 images for training and the remaining 9,000 for testing.

\noindent $\bullet$ \textbf{ShanghaiTech}~\cite{zhang2016single}\footnote{https://pan.baidu.com/s/1nuAYslz} includes two parts, i.e., Part\_A and Part\_B, with a total number of 1,198 images. The images of Part\_A are randomly crawled from the Internet, which are across diverse scenes and largely varied densities. Part\_A has 482 images, of which 300 are served as training set and the remaining 182 for testing. The images of Part\_B are taken from the metropolis in Shanghai, which consists of 400 images for training and 316 for testing.

\noindent $\bullet$ \textbf{UCF-QNRF}~\cite{idrees2018composition}\footnote{https://www.crcv.ucf.edu/data/ucf-qnrf/} is a recently released large and challenging dataset, which has a wide range of image resolutions, counts, scale variations and diversely density distribution. The images of this dataset are crawled from Flickr, Web Search and Hajj footage, containing 1,535 images with over 125 million point annotations, where 1,201 images are used for training and the remaining 334 images for testing.

\noindent $\bullet$ \textbf{UCF\_CC\_50}~\cite{idrees2013multi}\footnote{https://github.com/davideverona/deep-crowd-counting crowdnet} is composed of 50 images with various resolutions. The dataset is small-scale yet challenging since the average count is up to 1,280. Following~\cite{idrees2013multi}, five-fold cross-validation is performed to obtain the final test result.

\textbf{Evaluation protocol:} Two most widely used evaluation metrics, i.e., Mean Average Error (MAE) and Root Mean Squared Error (RMSE), are employed to evaluate the performance of the proposed method. The two metrics are defined as follows:
\begin{equation}
MAE = \frac{1}{K}\sum\limits_{i = 1}^K {\left| {\hat{C}_i - C_i} \right|}
\end{equation}

\begin{equation}
RMSE = \sqrt {\frac{1}{K}\sum\limits_{i = 1}^K {{{\left| {\hat{C}_i - C_i} \right|}^2}} }
\end{equation}
where $K$ is the number of test images, ${\hat{C}_i}$ denotes the predicted count and ${C_i}$ indicates the ground truth count for the $i$-th image, respectively.
\subsection{Implementation details}
We implement our proposed PSGCNet in PyTorch and train it in an end-to-end manner. All the experiments are conducted on one NVIDIA 2080Ti GPU. A truncated VGG19~\cite{simonyan2015very} pre-trained on ImageNet~\cite{krizhevsky2012imagenet} is taken as the backbone, with the fully connected layers and the last pooling layer removed. During training, the initial learning rate is 1e-5, and Adam optimizer is used. For better training and avoiding overfitting, random crop and horizontal flipping are applied for augmentation. Specifically, the crop size is $256\times256$ for RSOC\_building datasets, ShanghaiTech Part\_A, and UCF\_CC\_50, and $512\times512$ for RSOC\_small-vehicle, RSOC\_large-vehicle, RSOC\_ship, CARPK, PUCPR+, DroneCrowd, ShanghaiTech Part\_B and UCF\_QNRF, since they have large image sizes. In addition, for all the datasets, 10\% of the images are randomly sampled for validation from each training set. The batch size is set 1 for all the datasets.

\begin{figure*}[t]
	\centering
	\includegraphics[width=0.9 \linewidth]{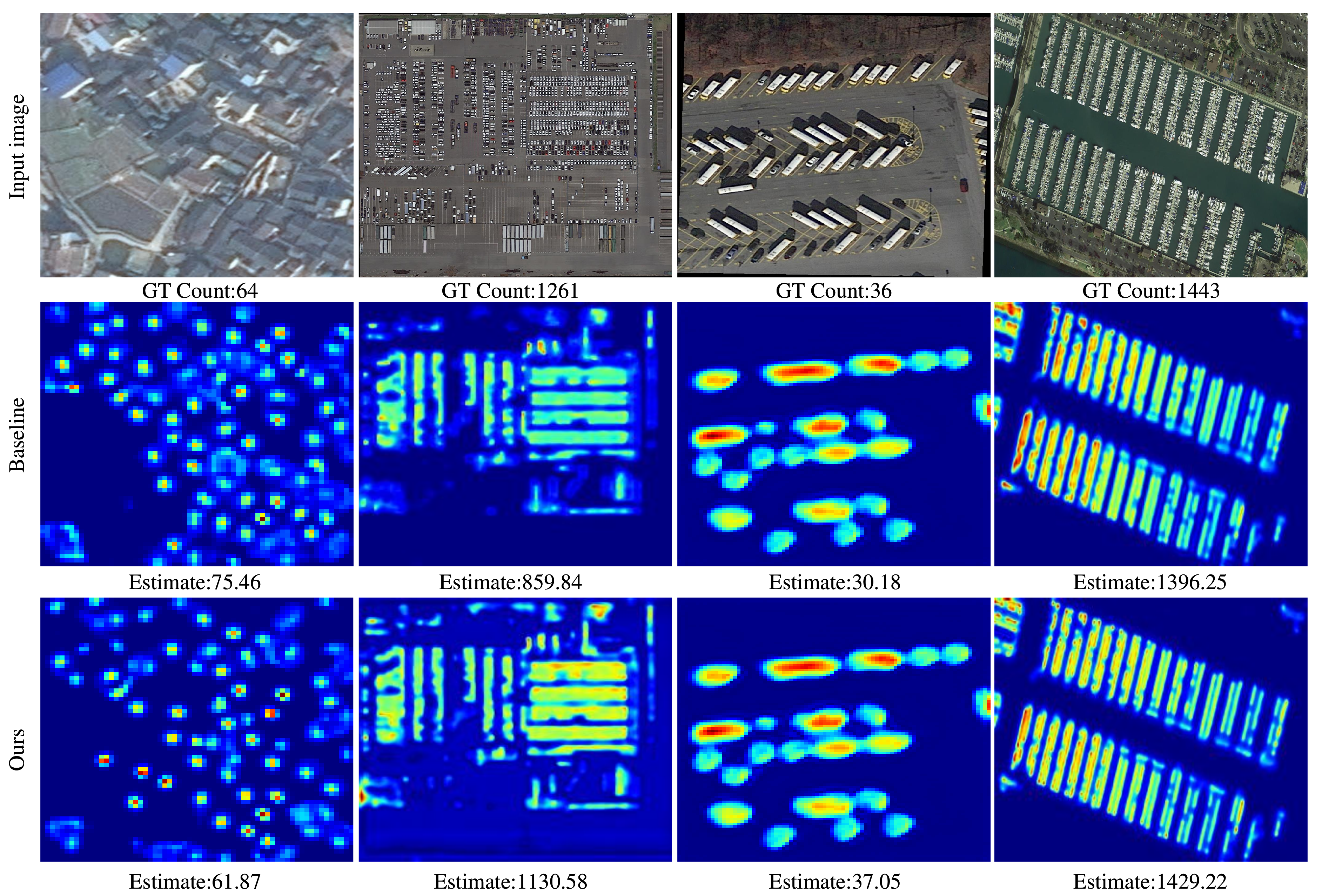}
	\caption{Density maps generated by Baseline (the middle row) and our method (the bottom row). The ground truth and estimated count are put at the bottom of each image. Compared with the baseline, our proposed model can obtain more accurate estimations across diverse scenarios.}
	\label{fig:denrsoc}
\end{figure*}

\begin{figure}[t]
	\centering
	\includegraphics[width=1.0 \linewidth]{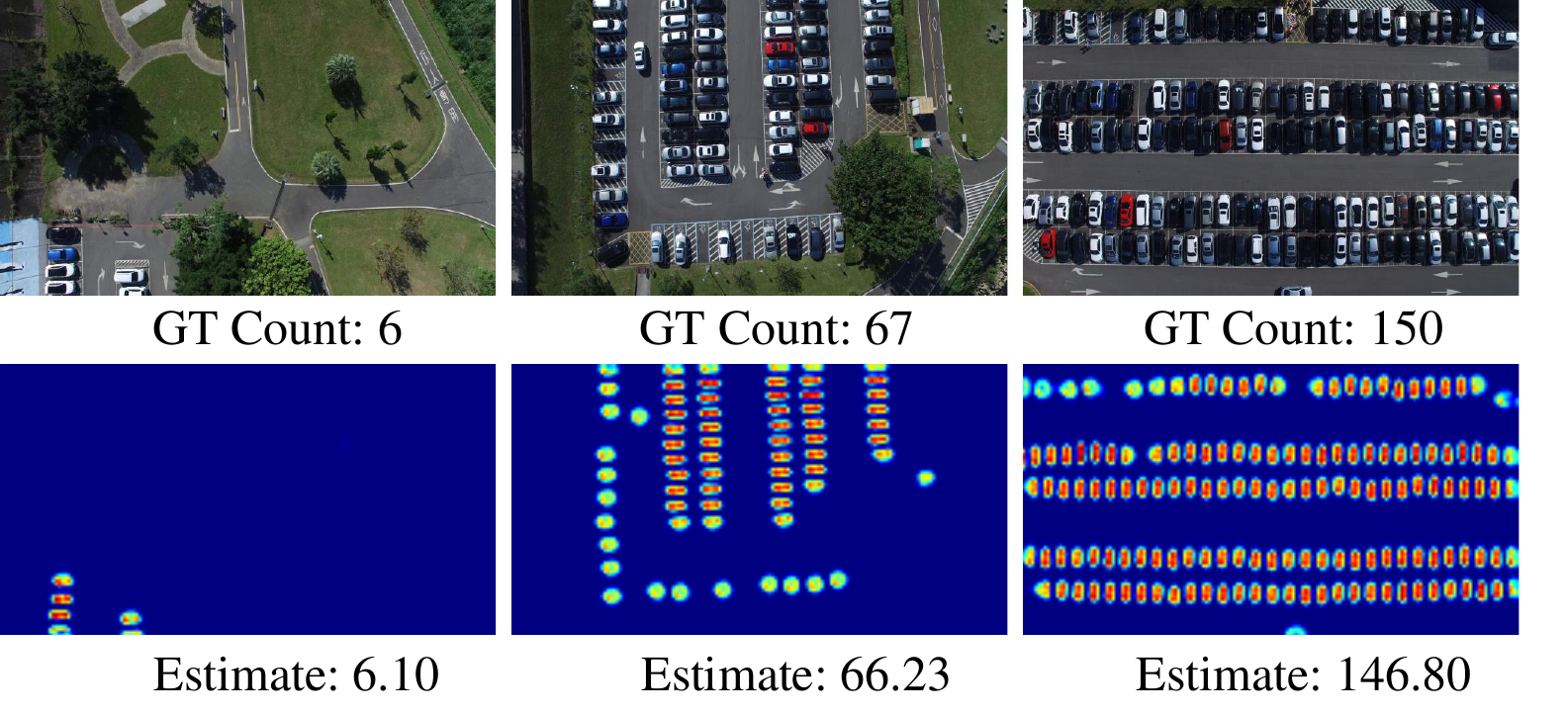}
	\caption{Visualization results on CARPK dataset. The top row shows the original image and the ground truth counts. The bottom one shows the density maps generated by our proposed method and the estimated counts.}
	\label{fig:CARPK}
\end{figure}

\begin{figure*}[t]
	\centering
	\includegraphics[width=1.0 \linewidth]{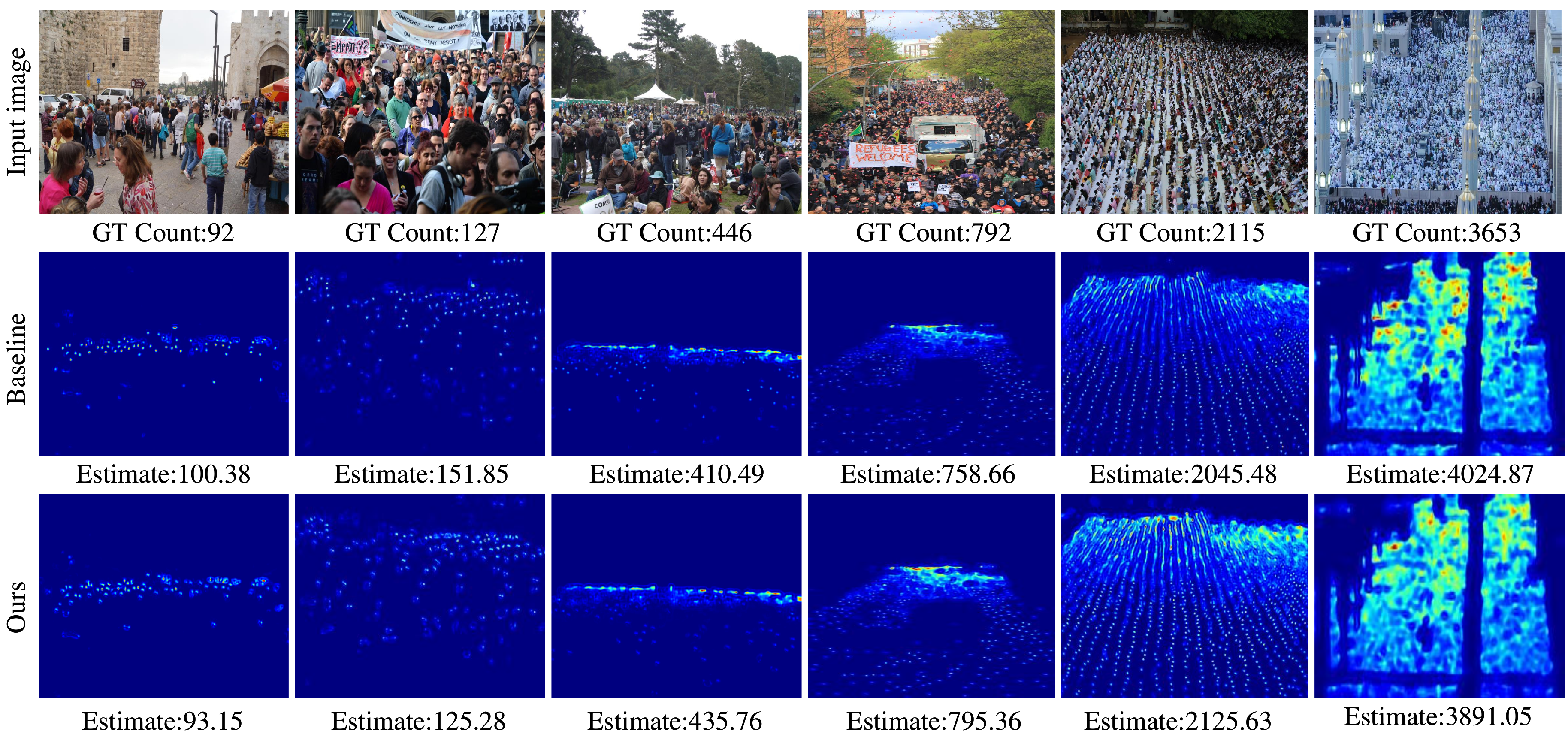}
	\caption{Density maps generated by Baseline (the middle row) and our method (the bottom row). The ground truth and estimated count are put at the bottom of each image. Compared with the baseline, our proposed model can obtain more accurate estimations from sparse to highly congested scenes.}
	\label{fig:visualziation}
\end{figure*}

\begin{table}[!t]
\caption{Different settings on RSOC\_building dataset.}
 \begin{center}
 \begin{tabular}{lll|ll}
\toprule
  Baseline & PSM &GCM & MAE & RMSE \\
\midrule
  \checkmark & -- & -- &11.51 &15.96   \\
  \checkmark  & \checkmark &-- &9.06 &13.24  \\
  \checkmark  &--  &\checkmark &8.64 &12.08  \\
  \checkmark &\checkmark &\checkmark &7.54 &10.52 \\
  \bottomrule
\end{tabular}
\end{center}
\label{tab:ablation}
\end{table}

\subsection{Ablation studies}
To validate the effectiveness of each module of our approach, we conduct ablation studies on RSOC\_building dataset. The baseline method is BL~\cite{ma2019bayesian}. The specific settings are shown in Table~\ref{tab:ablation}.

\noindent $\bullet$ \textbf{Effect of PSM and GCM.}
From Table~\ref{tab:ablation} we can observe that when PSM is introduced, the performance can achieve a significant improvement. Specifically, there are relative improvements of 21.28\% and 17.04\% w.r.t MAE and RMSE, demonstrating the robustness of the proposed PSM to the problem of large scale variation. To validate the robustness of the model to the complex background interference, we adopt a global context module. From Table~\ref{tab:ablation} we can find that the GCM can boost the baseline method with a considerable elevation. In particular, the performance will gain by 24.93\% and 24.31\% w.r.t. MAE and RMSE, which proves that it has made a significant impact on highlighting objects parts while diminishing background noise.

\noindent $\bullet$ \textbf{Effect of the hyperparameter $\lambda$.}
To verify the effectiveness of the proposed BCL loss function, we conduct experiments under the condition of different $\lambda$. As can be observed from Table~\ref{tab:lambda}, when $\lambda=0.1$, we can obtain the best performance.

\begin{table}[!t]
 \caption{Impacts of different $\lambda$ on RSOC\_building dataset.}
 \begin{center}
 \begin{tabular}{|l|c|c|}
  \hline
   & MAE & RMSE \\ \hline
  $\lambda$=0 &8.18 &12.61 \\
  $\lambda$=10 &8.36 &12.92 \\
  $\lambda$=1 &8.02 &11.96 \\
  $\lambda$=0.1 &\textbf{7.54} &\textbf{10.52} \\
  $\lambda$=0.01 &7.88 &11.02 \\
  $\lambda$=0.001 &7.94 &11.46 \\
  \hline
\end{tabular}
\end{center}
\label{tab:lambda}
\end{table}

\noindent $\bullet$ \textbf{Different backbones.}
Our proposed modules and loss function can be readily applied to any network structures to improve the performance. Here we apply them to VGG-19 and VGG-16 backbones and perform comparisons with the L2 loss-based models. The quantitative results from Table~\ref{table:backbone} show that our method achieves better performance compared with the naive L2-based modes by a considerable margin.

\begin{table}[htbp]
  \vspace{-0.3cm}
  \caption{Performances of using different backbones on RSOC\_building dataset.}
	\begin{center}
		\begin{tabular}{l|cc|cc|}
			\hline
			\multirow{2}{*}{\backslashbox{Methods}{Backbones}} & \multicolumn{2}{c|}{VGG-19}& \multicolumn{2}{c|}{VGG-16}\\
			\cline{2-5}
			&MAE &RMSE &MAE &RMSE \\
			\hline
            L2-loss &9.08 &12.48 &9.94 &14.04 \\ \hline
            PSGCNet(Ours) &\textbf{7.54} &\textbf{10.52} &\textbf{8.84 } &\textbf{12.18} \\
        \hline
		\end{tabular}
	\end{center}
\label{table:backbone}
\end{table}

\subsection{Comparisons on RSOC dataset}
We compare our approach with state-of-the-art methods and show results in Table~\ref{table:comparison_RSOC} and visualize some representative density maps in Fig.~\ref{fig:denrsoc}. Our model achieves substantial improvements on all the four subsets. Specifically, we improve the baseline with improvements of 34.49\%, 6.76\%, 17.85\% and 11.01\% on building, small-vehicle, large-vehicle and ship subsets in terms of MAE, respectively, indicating that our proposed method has a strong counting performance.

\begin{table*}[htbp!]
  \vspace{-0.3cm}
  \caption{Performance comparison on RSOC~\cite{gao2020counting} dataset.}
	\begin{center}
		\resizebox{\textwidth}{!}{
		\begin{tabular}{l|l|cc|cc|cc|cc}
			\hline
			\multirow{2}{*}{\backslashbox{Methods}{Datasets}} &\multirow{2}{*}{Year$\&$Venue} & \multicolumn{2}{c|}{RSOC\_Building}& \multicolumn{2}{c|}{RSOC\_Small-vehicle}&\multicolumn{2}{c|}{RSOC\_Large-vehicle}&\multicolumn{2}{c}{RSOC\_Ship}\\
			\cline{3-10}
			& &MAE &RMSE &MAE &RMSE &MAE &RMSE &MAE &RMSE  \\
			\hline
            MCNN~\cite{zhang2016single} &2016 CVPR &13.65 &16.56 &488.65 &1317.44 &36.56 &55.55 &263.91 &412.30 \\
            CMTL~\cite{sindagi2017cnn} &2017 AVSS &12.78 &15.99 &490.53 &1321.11 &61.02 &78.25 &251.17 &403.07 \\
            CSRNet~\cite{li2018csrnet} &2018 CVPR &8.00 &11.78 &443.72 &1252.22 &34.10 &46.42 &240.01 &394.81 \\
            SANet~\cite{cao2018scale} &2018 ECCV &29.01 &32.96 &497.22 &1276.66 &62.78 &79.65 &302.37 &436.91 \\
            SFCN~\cite{wang2019learning} &2019 CVPR &8.94 &12.87 &440.70 &1248.27 &33.93 &49.74 &240.16 &394.81 \\
            SPN~\cite{chen2019scale} &2019 WACV &7.74 &11.48 &445.16 &1252.92 &36.21 &50.65 &241.43 &392.88 \\
            SCAR~\cite{gao2019scar} &2019 NC &26.90 &31.35 &497.22 &1276.65 &62.78 &79.64 &302.37 &436.92 \\
            CAN~\cite{liu2019context-aware} &2019 CVPR &9.12 &13.38 &457.36 &1260.39 &34.56 &49.63 &282.69 &423.44 \\
            SFANet~\cite{zhu2019dual} &2019 CVPR &8.18 &11.75 &435.29 &1284.15 &29.04 &47.01 &201.61 &332.87 \\
             BL~\cite{ma2019bayesian} &2019 ICCV &11.51 &15.96 &168.62 &280.50 &13.39 &35.24 &84.18 &136.21\\
            ASPDNet~\cite{gao2020counting,gao2020counting_2} &2020 ICASSP/TGRS &7.59 &10.66 &433.23 &1238.61 &18.76 &31.06 &193.83 &318.95 \\
            PSGCNet(Ours) &-- &\textbf{7.54} &\textbf{10.52} &\textbf{157.55} &\textbf{245.31} &\textbf{11.00} &\textbf{17.65} &\textbf{74.91} &\textbf{112.11} \\
        \hline
		\end{tabular}}
	\end{center}
\label{table:comparison_RSOC}
\end{table*}

\begin{table}[htbp]
  \vspace{-0.3cm}
  \caption{Performance comparison on CARPK~\cite{hsieh2017drone} and PUCPR+~\cite{hsieh2017drone} dataset. ``*" indicates that the method has been fine-tuned on PUCPR+ dataset.}
	\begin{center}
		\begin{tabular}{l|cc|cc|}
			\hline
			\multirow{2}{*}{\backslashbox{Methods}{Datasets}} & \multicolumn{2}{c|}{CARPK~\cite{hsieh2017drone}}& \multicolumn{2}{c|}{PUCPR+~\cite{hsieh2017drone}}\\
			\cline{2-5}
			&MAE &RMSE &MAE &RMSE \\
			\hline
            YOLO~\cite{redmon2016you} &102.89 &110.02 &156.72 &200.54 \\ \hline
            *YOLO~\cite{redmon2016you} &48.89 &57.55 &156.00 &200.42 \\ \hline
            Faster RCNN~\cite{ren2016faster} &103.48 &110.64 &156.76 &200.59 \\ \hline
            *Faster RCNN~\cite{ren2016faster} &24.32 &37.62 &39.88 &47.67  \\ \hline
            One-Look Regression~\cite{mundhenk2016large} &59.46 &66.84 &21.88 &36.73  \\ \hline
            SSD~\cite{liu2016ssd} &37.33 &42.32 &119.24 &132.22 \\ \hline
            YOLO9000~\cite{redmon2017yolo9000} &38.59 &43.18 &97.96 &133.25 \\ \hline
            LPN~\cite{hsieh2017drone} &23.80 &36.79 &22.76 &34.46 \\ \hline
            RetinaNet~\cite{lin2017focal} &16.62 &22.30 &24.58 &33.12 \\ \hline
            LEP~\cite{stahl2018divide} &51.83 &-- &15.17 &-- \\ \hline
            MCNN~\cite{zhang2016single} &39.10 &43.30 &21.86 &29.53  \\ \hline
            CSRNet~\cite{li2018csrnet}  &11.48 &13.32 &8.65 &10.24 \\ \hline
            BL~\cite{ma2019bayesian} &9.58 &11.38 &6.54 &8.13 \\ \hline
            PSGCNet(Ours) &\textbf{8.15} &\textbf{10.46} &\textbf{5.24} &\textbf{7.36} \\
        \hline
		\end{tabular}
	\end{center}
\label{table:comparison_CARPK}
\end{table}

\subsection{Comparisons on CARPK and PUCPR+ dataset}
Table~\ref{table:comparison_CARPK} reports the MAE and RMSE results on two car counting datasets, i.e., CARPK and PUCPR+~\cite{hsieh2017drone}. We compare our proposed approach with state-of-the-art car counting methods including detection-based counting methods (YOLO~\cite{redmon2016you}, Faster RCNN~\cite{ren2016faster}, LPN~\cite{hsieh2017drone}, SSD~\cite{liu2016ssd}, YOLO9000~\cite{redmon2017yolo9000}, RetinaNet~\cite{lin2017focal}, and LEP~\cite{stahl2018divide}), a regression-based counting method (One-Look Regression~\cite{mundhenk2016large}), and density map estimation based methods (MCNN~\cite{zhang2016single}, CSRNet~\cite{li2018csrnet}, and BL~\cite{ma2019bayesian}). The results reveal that our method consistently performs better than the comparative methods, which demonstrate the superiority of our method both in sparse and congested scenarios. Specifically, compared with several outstanding object detectors such as Faster RCNN~\cite{ren2016faster} and YOLO~\cite{redmon2016you}, our proposed method surpasses them by a large margin. Moreover, compared with One-Look Regression~\cite{mundhenk2016large}, our approach shows better performance. We conjecture that it may be uncontrollable when regressing the count directly. Furthermore, compared with the density map estimation methods, i.e., MCNN~\cite{zhang2016single}, CSRNet~\cite{li2018csrnet}, and BL~\cite{ma2019bayesian}, our proposed method still obtains the highest count scores. We visualize some qualitative results in Fig.~\ref{fig:CARPK}. It demonstrates that the proposed method not only performs a better counting performance, but also shows strong localization ability.

\subsection{Comparisons on DroneCrowd dataset}
We also evaluate our method on a more challenging dataset, called DroneCrowd~\cite{wen2021detection}. Table~\ref{table:comparison_DroneCrowd} lists the counting results w.r.t MAE and RMSE, PSGCNet achieves comparable performance when compared with the state of the art methods. To further analyse the results, we also report the performance on several subsets according to three video-level attributes, i.e., two categories of scales including \emph{Large} (the diameter of objects $\textgreater$ 15 pixels) and \emph{Small} (the diameter of objects $\le$ 15 pixels), three categories of illumination conditions including \emph{Cloudy}, \emph{Sunny}, and \emph{Night}, two density levels including \emph{Crowded} (with the number of objects in each frame larger than 150) and \emph{Sparse} (with the number of objects in each frames less than 150). From the performance of subsets, we can find that our proposed method performs well in the \emph{Cloudy}, \emph{Sunny}, and \emph{Crowded} subsets, degrades in the \emph{Night} subset, this may be attributed to extremely low illumination and severe class imbalance. In particular, STNNet~\cite{wen2021detection} performs the best across the whole dataset. It is a multi-task learning model to jointly solve density map estimation, localization and tracking. The method also leverages both spatial and temporal information, in which a neighboring context loss is applied to capture relations among neighboring targets in consecutive frames. Even so, our proposed model achieves a comparatively good performance and even surpasses it in the \emph{Sunny} and \emph{Crowded} subsets.

\begin{table*}[htbp]
  \vspace{-0.3cm}
  \caption{Performance comparison on DroneCrowd dataset~\cite{wen2021detection}.}
	\begin{center}
		\resizebox{\textwidth}{!}{
		\begin{tabular}{l|c||cc||cc|cc||cc|cc|cc||cc|cc}
			\hline
			\multirow{2}{*}{Methods} &\multirow{2}{*}{Speed (FPS)} &\multicolumn{2}{c||}{Overall} &\multicolumn{2}{c|}{Large} &\multicolumn{2}{c||}{Small} &\multicolumn{2}{c|}{Cloudy} &\multicolumn{2}{c|}{Sunny} &\multicolumn{2}{c||}{Night} &\multicolumn{2}{c|}{Crowded} &\multicolumn{2}{c}{Sparse}\\
			\cline{3-18}
			& &MAE &RMSE &MAE &RMSE &MAE &RMSE &MAE &RMSE &MAE &RMSE &MAE &RMSE &MAE &RMSE &MAE &RMSE \\
			\hline
            MCNN~\cite{zhang2016single} &\textbf{28.98} &34.7 &42.5 &36.8 &44.1 &31.7 &40.1 &21.0 &27.5 &39.0 &43.9 &67.2 &68.7 &29.5 &35.3 &37.7 &46.2 \\ \hline
            CMTL~\cite{sindagi2017cnn} &2.31 &56.7 &65.9 &53.5 &63.2 &61.5 &69.7 &59.5 &66.9 &56.6 &67.8 &48.2 &58.3 &81.6 &88.7 &42.2 &47.9 \\ \hline
            MSCNN~\cite{zeng2017multi} &1.76 &58.0 &75.2 &58.4 &77.9 &57.5 &71.1 &64.5 &85.8 &53.8 &65.5 &46.8 &57.3 &91.4 &106.4 &38.7 &48.8\\ \hline
            LCFCN~\cite{laradji2018blobs} &3.08 &136.9 &150.6 &126.3 &140.3 &152.8 &164.8 &147.1 &160.3 &137.1 &151.7 &105.6 &113.8 &208.5 &211.1 &95.4 &110.0 \\ \hline
            SwitchCNN~\cite{sam2017switching} &0.01 &66.5 &77.8 &61.5 &74.2 &74.0 &83.0 &56.0 &63.4 &69.0 &80.9 &92.8 &105.8 &67.7 &79.8 &65.7 &76.7 \\ \hline
            ACSCP~\cite{shen2018crowd} &1.58 &48.1 &60.2 &57.0 &70.6 &34.8 &39.7 &42.5 &46.4 &37.3 &44.3 &86.6 &106.6 &36.0 &41.9 &55.1 &68.5 \\ \hline
            AMDCN~\cite{deb2018aggregated} &0.16 &165.6 &167.7 &166.7 &168.9 &163.8 &165.9 &160.5 &162.3 &174.8 &177.1 &162.3 &164.3 &165.5 &167.7 &165.6 &167.8 \\ \hline
            StackPooling~\cite{huang2018stacked} &0.73 &68.8 &77.2 &68.7 &77.1 &68.8 &77.3 &66.5 &75.9 &74.0 &83.4 &65.2 &67.4 &95.7 &101.1 &53.1 &59.1 \\ \hline
            DA-Net~\cite{zou2018net} &2.52 &36.5 &47.3 &41.5 &54.7 &28.9 &33.1 &45.4 &58.6 &26.5 &31.3 &29.5 &34.0 &56.5 &68.3 &24.9 &28.7 \\ \hline
            CSRNet~\cite{li2018csrnet} &3.92 &19.8 &25.6 &17.8 &25.4 &22.9 &25.8 &12.8 &16.6 &19.1 &22.5 &42.3 &45.8 &20.2 &24.0 &19.6 &26.5 \\ \hline
            CAN~\cite{liu2019context-aware} &7.12 &22.1 &33.4 &18.9 &26.7 &26.9 &41.5 &\textbf{11.2} &\textbf{14.9} &14.8 &17.5 &69.4 &73.6 &\textbf{14.4} &\textbf{17.9} &26.6 &39.7 \\ \hline
            DM-Count~\cite{wang2020DMCount} &10.04 &18.4 &27.0 &19.2 &29.6 &17.2 &22.4 &11.4 &16.3 &\textbf{12.6} &\textbf{15.2} &51.1 &55.7 &17.6 &21.8 &18.9 &29.6 \\ \hline
            STNNet~\cite{wen2021detection} &3.41 &\textbf{15.8} &\textbf{18.7} &\textbf{16.0} &\textbf{18.4} &\textbf{15.6} &\textbf{19.2} &14.1 &17.2 &19.9 &22.5 &\textbf{12.9} &\textbf{14.4} &18.5 &21.6 &\textbf{14.3} &\textbf{16.9} \\ \hline
            PSGCNet(Ours) &6.79 &24.7 &31.9 &24.5 &32.8 &24.0 &31.3 &18.8 &21.6 &15.4 &18.6 &47.5 &54.5 &18.2 &21.2 &18.6 &21.2 \\ \hline
		\end{tabular}}
	\end{center}
\label{table:comparison_DroneCrowd}
\end{table*}

\begin{table*}[ht!]
  \caption{Comparisons of our proposed PSGCNet with 15 state-of-the-art methods on four crowd counting datasets. The results of baseline is the reimplemented version with the pretrained weights at~\url{https://github.com/ZhihengCV/Bayesian-Crowd-Counting}.} 
	\begin{center}
		\resizebox{\textwidth}{!}{
		\begin{tabular}{l|l|ccc|ccc|ccc|ccc|c}
			\hline
			\multirow{2}{*}{\backslashbox{Methods}{Datasets}} &\multirow{2}{*}{Year$\&$Venue} &\multicolumn{3}{c|}{ShanghaiTech\_A}& \multicolumn{3}{c|}{ShanghaiTech\_B}&\multicolumn{3}{c|}{UCF\_QNRF} &\multicolumn{3}{c|}{UCF\_CC\_50} &\multicolumn{1}{c}{}\\
			\cline{3-15}
			& &MAE &RMSE &R. &MAE &RMSE &R. &MAE &RMSE &R. &MAE &RMSE &R. &avg.R. \\
			\hline
            MCNN~\cite{zhang2016single} &2016 CVPR &110.2 &173.2 &14.5 &26.4 &41.3 &13.5 &277 &426 &12 &377.6 &509.1 &16 &14 \\
            CMTL~\cite{sindagi2017cnn}  &2017 AVSS &101.3 &152.4 &13.5 &20.0 &31.1 &11.5 &252 &514 &12.5 &322.8 &397.9 &14 &12.9 \\
            Switching-CNN~\cite{sam2017switching} &2017 CVPR &90.4 &135 &12.5 &21.6 &33.4 &12.5 &228 &445 &11.5 &318.1 &439.2 &14.5 &12.8 \\
            CSRNet~\cite{li2018csrnet} &2018 CVPR &68.2 &115 &11 &10.6 &16 &9.5 &-- &-- &-- &266.1 &397.5 &11.5 &10.7 \\
            ACSCP~\cite{shen2018crowd} &2018 CVPR &75.7 &102.7 &9.5 &17.2 &27.4 &10.5 &-- &-- &-- &291.0 &404.6 &13.5 &11.2 \\
            SFCN~\cite{wang2019learning} &2019 CVPR &64.8 &107.5 &9.5 &7.6 &13.0 &6 &102.0 &171.4 &4.5 &214.2 &318.2 &7 &6.8 \\
            CAN~\cite{liu2019context-aware} &2019 CVPR &62.3 &100.0 &5 &7.8 &12.2 &5 &107 &183 &7.5 &212.2 &\textbf{243.7} &3 &5.1 \\
            TEDNet~\cite{jiang2019crowd} &2019 CVPR &64.2 &109.1 &9 &8.2 &12.8 &7 &113 &188 &9 &249.4 &354.5 &10 &8.8 \\
            BL~\cite{ma2019bayesian} &2019 ICCV &62.8 &101.8 &6 &7.7 &12.7 &5.5 &88.7 &154.8 &2 &229.3 &308.2 &6.5 &5 \\
            LS2M~\cite{xu2019learn} &2019 ICCV &64.2 &98.4 &5.5 &7.2 &11.1 &2.5 &104.7 &173.6 &5.5 &188.4 &315.3 &5 &4.6 \\
            DADNet~\cite{guo2019dadnet} &2019 MM &64.2 &99.9 &6 &8.8 &13.5 &8.5 &113.2 &189.4 &10 &285.5 &389.7 &11.5 &8.8 \\
            DUBNet~\cite{oh2020crowd} &2020 AAAI &64.6 &106.8 &8.5 &7.7 &12.5 &5 &105.6 &180.5 &6.5 &243.8 &329.3 &9 &7.3 \\
            HYGNN~\cite{luo2020hybrid} &2020 AAAI &60.2 &94.5 &2.5 &7.5 &12.7 &4.5 &100.8 &185.3 &6 &184.4 &270.1 &3.5 &4.1 \\
            SDANet~\cite{miao2020shallow} &2020 AAAI &63.6 &101.8 &6.5 &7.8 &10.2 &4 &-- &-- &-- &227.6 &316.4 &7 &5.8 \\
            ASNet~\cite{jiang2020attention} &2020 CVPR &57.78 &\textbf{90.13} &1.5 &-- &-- &-- &91.59 &159.71 &3 &\textbf{174.84} &251.63 &1.5 &2 \\ \hline
            Baseline &-- &64.5 &101.6 &-- &7.8 &13.5 &-- &98.6 &169.7 &-- &229.3 &308.2 &--&-- \\
            PSGCNet(Ours) & -- &\textbf{56.1} &95.6 &2 &\textbf{6.6} &\textbf{9.7} &1 &\textbf{86.3} &\textbf{149.5} &1 &181.3 &263.5 &2.5 &\textbf{1.6} \\
        \hline
		\end{tabular}}
	\end{center}
\label{table:comparison_crowd}
\end{table*}

\subsection{Comparisons on crowd counting datasets}
To further validate the generalization ability and robustness of the proposed model, we extend it on four widely used crowd counting datasets, the counting results are reported in Table~\ref{table:comparison_crowd}. It demonstrates that our proposed approach can achieve consistent improvements compared with 15 state-of-the-art methods~\cite{zhang2016single,sindagi2017cnn,sam2017switching,li2018csrnet,shen2018crowd,wang2019learning,liu2019context-aware,jiang2019crowd,liu2019crowd,
ma2019bayesian,xu2019learn,guo2019dadnet,oh2020crowd,luo2020hybrid,miao2020shallow,jiang2020attention}. Specifically, on ShanghaiTech dataset, our proposed model increases relative improvements of 12.4\%/5.9\% on Part\_A and 15.4\%/28.1\% on Part\_B, w.r.t, MAE/RMSE. Even on the more crowded UCF\_QNRF and UCF\_CC\_50, we still improve the baseline with relative improvements of 12.5\%/11.9\% and 20.9\%/14.5\% w.r.t MAE/RMSE. It indicates that our proposed method achieves superior performance not only for sparse but also highly congested crowd scenes.

In consideration of some methods that may perform well on one dataset however poorly on other ones, for fairness, we adopt the average ranking evaluation strategy~\cite{jiang2020attention} to make a comprehensive evaluation (denoted by avg. R. in Tabel~\ref{table:comparison_crowd}). The average ranking value is obtained by summing all ranks that one method gains to divide the number of datasets it utilizes. The lower value indicates a higher rank. Therefore, our proposed method obtains the best average ranking, which reveals its powerful ability to deal with the diverse crowd scenes.

We visualize some estimated density maps of the proposed method and the baseline in Fig.~\ref{fig:visualziation}, from which we can observe that our proposed method obtains more accurate estimations. Benefiting from the proposed PSM and GCM, our method can better reflect the scale variation of the pedestrians. Compared with the baseline method, our proposed model obtains more accurate estimations across diverse scenes from sparse to highly congested. Moreover, compared with baseline, our method obtains clearer density maps and shows stronger localization ability to a certain extent.


\section{Conclusion}
\label{sec:conclusion}
In this paper, we have presented a novel supervised learning framework for dense object counting in remote sensing images, named PSGCNet. Our PSGCNet is characterized by three components: 1) capturing multi-scale features with an effective pyramidal scale module; 2) alleviating the interferences of complex background with a lightweight global context module, and 3) a reliable supervision manner combined with Bayesian loss and counting loss, which is utilized to train the network and learn the count expectation at each annotation point. Extensive experiments on four remote sensing object counting  datasets demonstrate the effectiveness and superiority of the proposed approach. Moreover, extension experiments on four widely used crowd counting benchmark datasets further validate the generalization ability and robustness of the model.
\bibliographystyle{IEEEtran}
\small
\bibliography{IEEEabrv,references}

\end{document}